\documentclass[sigconf]{acmart} 
\usepackage{tabularx}
\usepackage{amsmath}
\usepackage{soul,color}
\usepackage{multirow}
\usepackage{tipa}
\usepackage{amsthm}
\theoremstyle{definition}
\newtheorem{definition}{Definition}
\usepackage{bbold}
\usepackage{mathrsfs}
\usepackage{booktabs} 
\usepackage{enumitem} 
\usepackage{array}
\usepackage{tikz}

\newcolumntype{M}[1]{>{\centering\arraybackslash}m{#1}}

\newcommand\red[1]{\textcolor{red}{#1}}

\newcommand\blue[1]{\textcolor{blue}{#1}}

\newcommand\magenta[1]{\textcolor{magenta}{#1}}

\newcommand{\Poincare}{Poincar\'e\ }
\newcommand{\expmap}{exponential\ }
\newcommand{\logmap}{logarithmic\ }

\newcommand{\ours}{\texttt{D-HYPR}\ }
\newcommand{\ourseos}{\texttt{D-HYPR}}

\newcommand{\norm}[1]{\left\lVert#1\right\rVert}

\copyrightyear{2022}
\acmYear{2022}
\setcopyright{acmlicensed}
\acmConference[CIKM '22]{Proceedings of the 31st ACM International Conference on Information and Knowledge Management}{October 17--21, 2022}{Atlanta, GA, USA} \acmBooktitle{Proceedings of the 31st ACM International Conference on Information and Knowledge Management (CIKM '22), October 17--21, 2022, Atlanta, GA, USA}
\acmPrice{15.00}
\acmDOI{10.1145/3511808.3557344} \acmISBN{978-1-4503-9236-5/22/10}

\begin{document}

\title{\ourseos: Harnessing Neighborhood Modeling and Asymmetry Preservation for Digraph Representation Learning}

\author{Honglu Zhou}
\affiliation{%
  \institution{Rutgers University}
  \streetaddress{10 Frelinghuysen Rd}
  \city{Piscataway, NJ}
  \country{US}}
\email{honglu.zhou@rutgers.edu}

\author{Advith Chegu}
\affiliation{%
  \institution{Rutgers University}
  \streetaddress{10 Frelinghuysen Rd}
  \city{Piscataway, NJ}
  \country{US}}
\email{ac1771@rutgers.edu}

\author{Samuel S. Sohn}
\affiliation{%
  \institution{Rutgers University}
  \streetaddress{10 Frelinghuysen Rd}
  \city{Piscataway, NJ}
  \country{US}}
\email{sss286@cs.rutgers.edu}

\author{Zuohui Fu}
\affiliation{%
  \institution{Rutgers University}
  \streetaddress{10 Frelinghuysen Rd}
  \city{Piscataway, NJ}
  \country{US}}
\email{zuohui.fu@rutgers.edu}

\author{Gerard de Melo}
 \affiliation{
   \institution{HPI / University of Potsdam} 
  \city{Potsdam}
  \country{Germany}
    }
\email{gerard.demelo@hpi.de}

\author{Mubbasir Kapadia}
\affiliation{%
  \institution{Rutgers University}
  \streetaddress{10 Frelinghuysen Rd}
  \city{Piscataway, NJ}
  \country{US}}
\email{mk1353@cs.rutgers.edu}

\renewcommand{\shortauthors}{Honglu Zhou et al.}

\begin{abstract}
Digraph Representation Learning (DRL) aims to learn representations for directed homogeneous graphs (digraphs). 
Prior work in DRL is largely constrained (e.g., limited to directed acyclic graphs), or has
 poor generalizability across tasks (e.g., evaluated solely on one task). 
Most Graph Neural Networks (GNNs) exhibit poor performance on digraphs due to the neglect of 
modeling neighborhoods and 
preserving asymmetry. 
In this paper, we address these notable challenges 
by leveraging \emph{hyperbolic collaborative learning} from 
multi-ordered and partitioned neighborhoods,
and
regularizers inspired by \emph{socio-psychological factors}. 
Our resulting formalism, Digraph Hyperbolic Networks 
(\ourseos) -- albeit conceptually simple -- generalizes to digraphs where cycles and non-transitive relations are common, and is applicable to multiple downstream tasks including node classification, link presence prediction, and link property prediction.
In order to assess the effectiveness of \ourseos, extensive evaluations were performed across $8$ real-world digraph datasets involving $21$ prior techniques.
\ours statistically significantly outperforms the current state of the art.
We release our code at \magenta{\url{https://github.com/hongluzhou/dhypr}}

\end{abstract}



\begin{teaserfigure}
  \includegraphics[width=\textwidth]{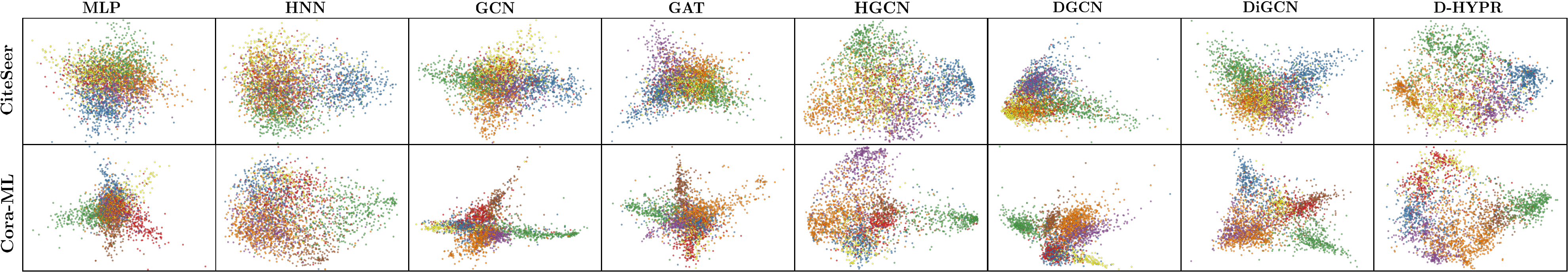}
\caption{Embedding space visualization using PCA projection.  Our method \ours leads to the best class separation.  Each dot represents a node, and colors reflect the ground truth class labels of nodes, which are best appreciated when zooming in. 
	}
  \label{fig:teaser}
\end{teaserfigure}

\maketitle

\section{Introduction}
\label{sec:intro}

\noindent Directionality is a fundamental characteristic inherent in a multitude of real-world graphs, including social networks, web page networks, and citation networks~\cite{hope}.
Digraph Representation Learning (DRL) aims to learn representations for directed homogeneous graphs (digraphs)~\cite{digcn,zhang2021magnet}. 
Early DRL techniques
include factorization-based approaches such as HOPE~\cite{hope} and ATP~\cite{atp}, and random walk-based approaches such as APP~\cite{app} and NERD~\cite{nerd}.
However, these methods do not scale to large digraphs, or are sensitive to outliers and noise.
In recent years, Graph Neural Networks (GNNs) have achieved immense success on a wide range of tasks~\cite{zhou2018graph}.
However, GNNs primarily aim at representation learning for \textit{undirected} graphs.
There are two notable challenges that hinder their effectiveness on digraphs.

\textit{ \textbf{Challenge 1: Neighborhood Modeling.}} The neighborhoods of a node may possess unique semantics. 
For instance, in social networks, in-neighbors are commonly known as followers, while out-neighbors are accounts that the user follows. In citation networks, in-neighbors of a node can be
existing works cited by a paper by the time the camera-ready version of the paper is submitted, whereas out-neighbor connections arise
subsequent to the paper
coming out.
Existing GNN techniques~\cite{vgae,gat,hgcn,gil}
transform digraphs to undirected graphs to enable running 
experiments, which simplifies the learning problem, or only consider the direct out-neighbors 
in the graph convolution. Thus, they lose characteristics of the original structure, resulting in
misleading message passing and ultimately subpar results on digraph-specific problems.

\textit{\textbf{Challenge 2: Asymmetry Preservation.}}
Due to the inherent symmetry of
popular
measures, 
such as the inner product or distance in the embedding space, which produces the same 
scores for the node pair $(i, j)$ and $(j, i)$,
inner-product- or distance-based 
learning objectives used by popular GNNs
are unsuitable for capturing
the asymmetric connection
probabilities for node pairs in digraphs~\cite{gravity}.
Applications based on link prediction
or graph topology learning
are particularly affected when models fail to preserve digraph structural asymmetry.

Recently, spectral-based DRL GNNs~\cite{zhang2021magnet,digcn,dgcn,ma2019spectral} have been proposed to address \textit{\textbf{Challenge 1}} with respect to modelling neighborhoods for digraphs.
However, the learned filters from these methods depend on the Laplacian eigenbasis, which is tied to a graph's structure~\cite{tong2021directed}. 
Models trained on a specific structure cannot be directly applied to graphs with different structures~\cite{wu2020comprehensive}.
Separately, to address \textit{\textbf{Challenge 2}}, approaches such as viewing directions of edges as a kind of edge feature~\cite{gong2019exploiting}, or
parametrizing the node pair likelihood function by a neural network~\cite{shi2019skeleton,battaglia2018relational}
have been proposed, but these techniques fail to consider \textit{\textbf{Challenge 1}}.
Moreover, prior DRL techniques are often constrained to directed acyclic graphs (DAGs)~\cite{thost2021directed,cones,suzuki2019hyperbolic}, are transductive~\cite{sim2021directed,dgcn,cones,suzuki2019hyperbolic}, or have poor generalizability across tasks. For example, some studies provide experimental evidence for a single task only -- e.g., link prediction as in~\cite{sim2021directed} or node classification as in~\cite{dgcn,ma2019spectral}.

We propose
\textbf{D}igraph \textbf{HYPER}bolic Networks
(\ourseos)  to fully address these limitations. 
To overcome \textit{\textbf{Challenge 1}}, 
\ours utilizes \emph{hyperbolic collaborative learning} from multi-ordered and partitioned
neighborhoods.
For \textit{\textbf{Challenge 2}}, \ours takes advantage of self-supervised learning, using asymmetry-preserving regularizers
supported by well-established  socio-psychological theories~\cite{birds,mitzenmacher2004brief}. Specifically:
\vspace{-10pt}
\begin{enumerate}[leftmargin=*]
    \item \ul{Neighborhood Modeling with Partitioned and Larger Receptive Fields}: by leveraging collaborative learning from multi-ordered four canonical types of  neighborhoods (Fig.~\ref{fig:model} (a)), \ours
     models the distinct node neighborhoods, and captures the local directed graph characteristics.
    
    \item \ul{Neighborhood Modeling with a non-Euclidean Space}: \ours learns node representations of real-world digraphs (which exhibit scale-free or hierarchical structures) in hyperbolic space to avoid
    distortion of node neighborhoods.
    
    \item \ul{Asymmetry Preservation with Regularizers}: motivated by two decomposed causes of link formation,  \textit{homophily}~\cite{birds} and \textit{preferential attachment}~\cite{mitzenmacher2004brief}, we employ two regularizers in training \ourseos, which are used in a self-supervised fashion to account for each of the two driving forces of link formation.
    These regularizers lead to
    performance gains in downstream tasks. 
    
    \item \ul{Flexibility due to Message-passing-based GNN Formalism}: \ours falls into the category of message-passing-based GNNs that capture both graph structure and semantics. \ours has the capability to inductively learn representations for \textit{general} digraphs that potentially contain cycles,  non-transitive relations, outliers, and noise.
\end{enumerate}

\noindent Our contributions are three-fold: \textbf{(1)} We propose \ours for DRL. \ours considers the unique node neighborhoods in digraphs with multi-scale neighborhood collaboration in hyperbolic space.
\ours respects asymmetric relationships of node-pairs, which is guided by sociopsychology-inspired regularizers. \textbf{(2)} We perform extensive benchmarking experiments across \textit{\textbf{8}} real-world digraph datasets. Our evaluation involves \textit{\textbf{4}} tasks and \textit{\textbf{21}} prior methods. Results 
demonstrate the significant superiority of \ours against the state of the art. 
\textbf{(3)} \ours generates meaningful embeddings in very low dimensionalities. This added benefit is desirable for large-scale real-world applications by efficiently saving space while preserving effectiveness.

\section{Related Work}
\label{sec:related}

\noindent \textbf{Graph Representation Learning (GRL).} GRL methods have evolved from matrix factorization~\cite{jiang2016dimensionality}, graph kernels~\cite{shervashidze2011weisfeiler}, and random walk-based transductive models~\cite{deepwalk}, into GNNs~\cite{gcn}, which have greatly surpassed these prior methods in numerous experiments.
Interested readers may refer to comprehensive reviews 
\cite{zhou2018graph,cai2018comprehensive,graphkernelsurvey,wu2020comprehensive} for further details. 
Current popular GRL approaches~\cite{beani2021directional,gcn,gat,hgcn,gil,hyperbolicgat,beani2021directional} 
have primarily considered \textit{undirected} homogeneous GRL.
Although certain recent GNNs can be applied to digraphs, e.g., the Graphormer~\cite{ying2021transformers} with its Transformer-based design~\cite{vaswani2017attention}, these techniques have been validated solely by experiments on undirected graphs~\cite{zhang2021magnet}, and are computationally impractical for large-scale digraphs.

\noindent \textbf{Directed Graph Embedding.} 
There are comparatively few studies that address DRL.
HOPE~\cite{hope} captures asymmetric transitivity
but depends on a low rank assumption of the input, and 
fails to generalize to a variety of tasks~\cite{nerd}.
APP~\cite{app} captures asymmetry by
relying on random walks.
ATP~\cite{atp} 
removes cycles in digraphs beforehand and then leverages factorization.
NERD~\cite{nerd} extracts a source and a target walk, and 
employs a shallow neural model. 
DGCN~\cite{dgcn}, DiGCN~\cite{digcn} and MagNet~\cite{zhang2021magnet} are recent GNNs that extend \textit{spectral-based} GCNs~\cite{gcn} to digraphs, but are tied to a graph’s Laplacian. 
DAGNN~\cite{thost2021directed} is proposed for DAGs by injecting a DAG-specific inductive bias---partial ordering---into the GNN design.

\noindent \textbf{Hyperbolic Embedding Learning.} 
Most non-Euclidean embedding techniques~\cite{hnn,nickel2017poincare,gu2018learning,nickel2018learning,law2020ultrahyperbolic,sim2021directed}
only account for the graph structure and do \textit{not} leverage node features. 
In contrast, we consider the \textit{general} DRL setting of seeking to capture \textit{both} digraph structure and attributes, and propose a message-passing-based GNN with an inductive learning capability.

HGCN~\cite{hgcn} and HGNN~\cite{hgnn} were proposed concurrently to generalize GNNs to take advantage of the strength of hyperbolic geometries.
Other hyperbolic GNNs include Constant Curvature GCNs~\cite{constantcurvature} that provide a mathematically grounded generalization of GCNs, HAT~\cite{hyperbolicgat} that studies hyperbolic GNN with an attention mechanism,  GIN~\cite{gil} that draws on both Euclidean and hyperbolic geometries, and so on~\cite{dai2021hyperbolic,chen2021fully,zhang2021lorentzian}. Our work is built upon these prior efforts on  hyperbolic GNNs for undirected graphs, to address challenges associated with digraphs.

\section{Preliminaries}
\label{sec:preli}

\begin{definition}
\textbf{\textit{Digraph Representation Learning}}~\cite{digcn,zhang2021magnet}.
Let $\mathcal{G}=(\mathcal{V}, \mathcal{E})$ be a homogeneous graph  with vertex set $\mathcal{V}$
and edge set $\mathcal{E}$.
Each edge $e\in \mathcal{E}$ is an ordered pair $e=(i,j)$ between vertices $i$ and $j$. The 
adjacency matrix of $\mathcal{G}$ can be denoted as $A=\{0,1\}^{|\mathcal{V}| \times |\mathcal{V}|}$. $\mathcal{G}$ is a digraph when $ {\exists}(i, j), A_{i,j}\neq A_{j, i}$.
\end{definition}
\vspace{-5pt}
\noindent Nodes are described by a feature matrix $X^{0, E} \in \mathbb{R}^{|\mathcal{V}| \times d}$, i.e., each node $i\in \mathcal{V}$ has a $d$-dimensional Euclidean feature $\mathbf{x}_{i}^{0, E}$.
The superscript $^E$ indicates that the vector
lies in a Euclidean space, while $^H$ denotes a hyperbolic vector. 
$0$ denotes the input layer. 
 
DRL is an effective and efficient solution for digraph analytics. The efficiency is achieved by converting the adjacency-matrix-based data into low-dimensional embeddings. 
Thus, the goal of DRL is to learn a mapping 
\vspace{-5pt}
\begin{equation}
\small
    f:\left(\mathcal{V}, \mathcal{E},\left(\mathbf{x}_{i}^{0, E}\right)_{i \in \mathcal{V}}\right) \rightarrow Z \in \mathbb{R}^{|\mathcal{V}| \times d^{\prime}}
\vspace{-3pt}
\end{equation}
that maps nodes to low-dimensional ($d^{\prime} \ll|\mathcal{V}|$) embedding vectors.
These should capture both structural and semantic information and be valuable
for downstream 
tasks.

\begin{definition} \textit{\textbf{The \Poincare Ball Model.}}\footnote{Our
method is compatible with other non-Euclidean embedding models.  }
The \Poincare ball model~\cite{hnn} $\left(\mathbb{D}_{c}^{n}, g^{c}\right)$  is defined by the
$n$-dimensional 
manifold $\mathbb{D}_{c}^{n}=\{x\in \mathbb{R}^{n}:$ $c\norm{\mathbf{x}}<1\}$ equipped with the Riemannian metric: $g_{\mathbf{x}}^{c}=\lambda_{\mathbf{x}}^{2} g^{E}$, where $\lambda_{\mathbf{x}}:=\frac{2}{1-c\norm{\mathbf{x}}^{2}}$, $g^{E}=\mathbf{I}_{n}$ is the Euclidean metric tensor, and  
$c > 0$ (we refer to $-c$ as the curvature). 
$\mathbb{D}_{c}^{n}$ is the open ball of radius $1 / \sqrt{c}$. 
The connections between hyperbolic space and tangent space are established by the \textit{\expmap map} $\exp _{\mathbf{x}}^{c}
: \mathcal{T}_{\mathbf{x}} \mathbb{D}_{c}^{n} \rightarrow \mathbb{D}_{c}^{n}$ and \textit{\logmap map} $\log _{\mathbf{x}}^{c}: \mathbb{D}_{c}^{n} \rightarrow \mathcal{T}_{\mathbf{x}} \mathbb{D}_{c}^{n}$:
\begin{equation}
\small
    \exp _{\mathbf{x}}^{c}(\mathbf{v})=\mathbf{x} \oplus_{c}\left(\tanh \left(\sqrt{c} \frac{\lambda_{\mathbf{x}}^{c}\norm{\mathbf{v}}}{2}\right) \frac{\mathbf{v}}{\sqrt{c}\norm{\mathbf{v}}}\right)
    \label{equa:expmap}
\end{equation}

\vspace{-10pt}
\begin{equation}
\small
    \log _{\mathbf{x}}^{c}(\mathbf{y})=\frac{2}{\sqrt{c} \lambda_{\mathbf{x}}^{c}} \tanh ^{-1}\left(\sqrt{c}\norm{-\mathbf{x} \oplus_{c} \mathbf{y}}\right) \frac{-\mathbf{x} \oplus_{c} \mathbf{y}}{\norm{-\mathbf{x} \oplus_{c} \mathbf{y}}}
    \label{equa:logmap}
\end{equation}
where $\mathbf{x}, \mathbf{y} \in \mathbb{D}_{c}^{n}$, $\mathbf{v} \in \mathcal{T}_{\mathbf{x}} \mathbb{D}_{c}^{n}$, and $\oplus_{c}$ denotes \textit{Möbius addition}, and
\begin{equation}
    \small
    \mathbf{x} \oplus_{c} \mathbf{y}:=\frac{\left(1+2 c\langle\mathbf{x}, \mathbf{y}\rangle+c\norm{\mathbf{y}}^{2}\right) \mathbf{x}+\left(1-c\norm{\mathbf{x}}^{2}\right) \mathbf{y}}{1+2 c\langle\mathbf{x}, \mathbf{y}\rangle+c^{2}\norm{\mathbf{x}}^{2}\norm{\mathbf{y}}^{2}}.
    \label{equa:mobius_add}
\end{equation}
The \textit{Möbius scalar multiplication} (Eq.~\ref{equa:scalar_mul}) and \textit{Möbius matrix multiplication} of $\mathbf{x} \in \mathbb{D}_{c}^{n} \backslash\{\mathbf{0}\}$ (Eq.~\ref{equa:matrix_mul}) are

\vspace{-5pt}
\begin{equation}
\small
    r \otimes_{c} \mathbf{x}:=\frac{1}{\sqrt{c}} \tanh \left(r \tanh ^{-1}(\sqrt{c}\norm{\mathbf{x}})\right) \frac{\mathbf{x}}{\norm{\mathbf{x}}}
    \label{equa:scalar_mul}
\end{equation}

\vspace{-10pt}
\begin{equation}
    \fontsize{8.5pt}{8.5pt}\selectfont
    M \otimes_{c} \mathbf{x}:=(1 / \sqrt{c}) \tanh \left(\frac{\norm{M \mathbf{x}}}{\norm{\mathbf{x}}} \tanh ^{-1}(\sqrt{c}\norm{\mathbf{x}})\right) \frac{M \mathbf{x}}{\norm{M \mathbf{x}}}
    \label{equa:matrix_mul}
\end{equation}
where $r \in \mathbb{R}$ and $M \in \mathbb{R}^{m \times n}$.
The induced distance function on $\left(\mathbb{D}_{c}^{n}, g^{c}\right)$ is given by 

\vspace{-6pt}
\begin{equation}
\small
    d_{\mathbb{D}_{c}^{n}}(\mathbf{x}, \mathbf{y})=(2 / \sqrt{c}) \tanh ^{-1}\left(\sqrt{c}\norm{\mathbf{-x} \oplus_{c} \mathbf{y}}\right)
    \label{equa:dist}
\end{equation}
\end{definition}

\noindent For a longer introduction of hyperbolic or non-Euclidean geometry, we refer readers to relevant previous work~\cite{bronstein2017geometric,chamberlain2017neural,hnn,hgcn,tradeoffs}.

\section{Methodology}
\label{sec:method}

Driven by the goal of addressing the challenge of Neighborhood Modeling and Asymmetry Preservation in digraphs, we propose \ours (Fig.~\ref{fig:model} (b)), which leverages hyperbolic collaborative learning from multi-ordered and partitioned neighborhoods, and self-supervised learning via asymmetry-preserving regularizers. 

Euclidean space does not provide the most powerful or meaningful geometrical representations when input data exhibits a highly complex non-Euclidean latent anatomy, as for instance for real-world digraphs with a scale-free or hierarchical structure~\cite{bronstein2017geometric,chamberlain2017neural}. 
As the volume of nodes grows exponentially with the distance from a central node, non-Euclidean geometry is more suitable than Euclidean for embedding such digraphs~\cite{tradeoffs,sim2021directed,law2020ultrahyperbolic}.
Hyperbolic embeddings can incur smaller data distortion for real-world digraphs, which leads to a better representation of the nodes' local neighborhoods. This motivates our investigation of utilizing hyperbolic GNNs over Euclidean counterparts as the backbone for DRL.

\subsection{Hyperbolic Embedding Learning}
\label{sec:hyperbolic_embed}

\begin{figure*}[t]
	\centering
	\includegraphics[scale=0.45]{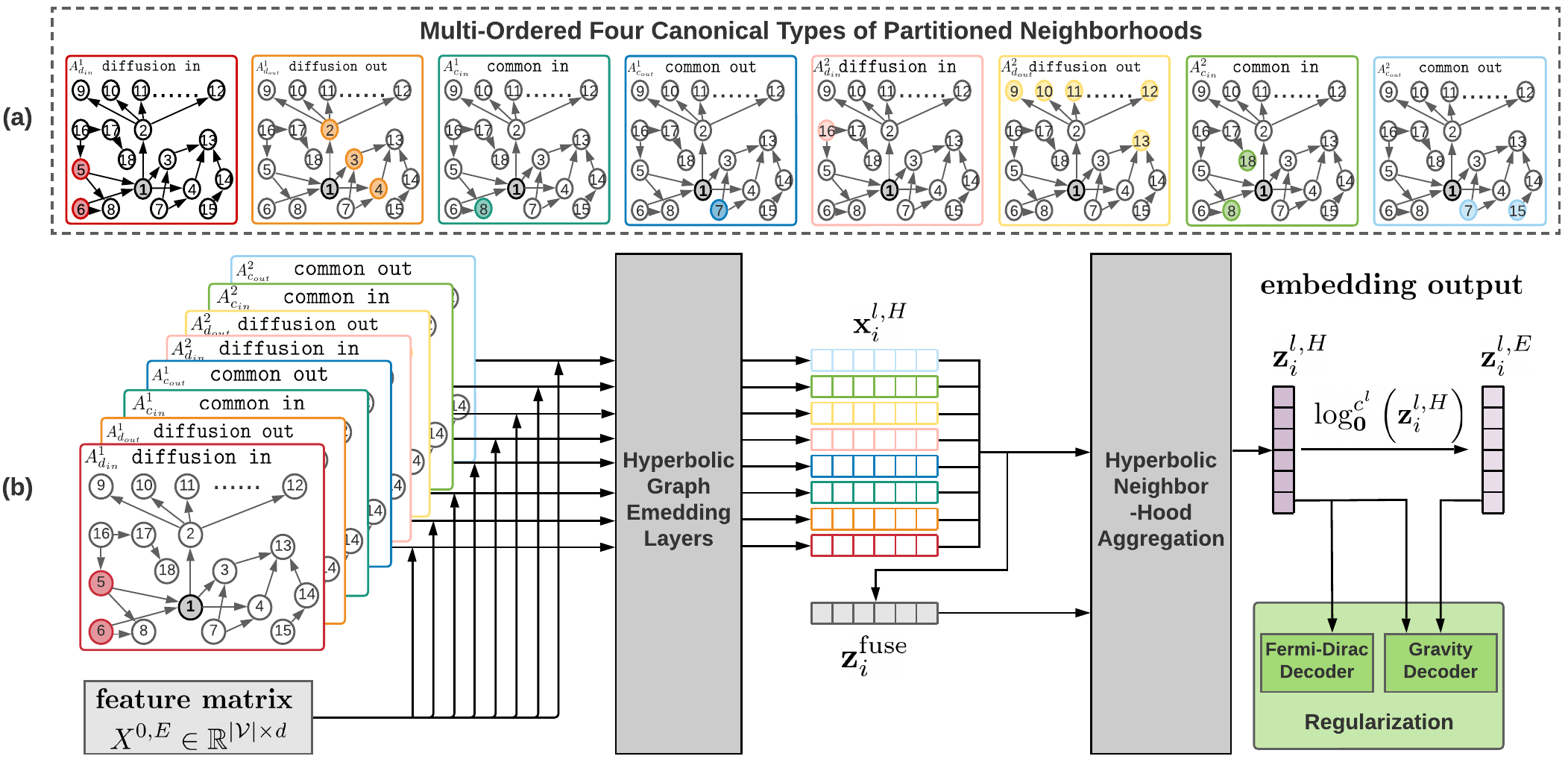}
	\caption{\textbf{(a) Multi-ordered and partitioned neighborhoods}. 
	We define four types of $k$-order proximity matrix (shown in the figure
	with $k=1,2$ with respect to node $1$) to incorporate the pertinent subsets of neighbors and multi-scale information.
	\textbf{(b) Methodology overview.} \ours learns a node representation from each 
	neighborhood in hyperbolic space
	with \textit{Hyperbolic Graph Embedding Layers}.
\textit{Hyperbolic Neighborhood Aggregation}
further enables a closer collaboration of neighborhoods. \ours respects asymmetric relationships of nodes with the hyperbolic Fermi-Dirac and Gravity regularizers.
}
	\label{fig:model}
\end{figure*}

To perform message passing in hyperbolic space, the general \emph{efficient} approach is to move basic operations of hyperbolic space to the tangent space~\cite{hyperbolicgat,gil}.
Given $\mathcal{G}$ and $\mathbf{x}_{i}^{0, E}$, we first obtain $\mathbf{x}_{i}^{0, H}$ by applying \expmap map $\exp _{\mathbf{0}}^{c^{0}}(\cdot)$ 
to map the Euclidean input feature $\mathbf{x}_{i}^{0, E}$ into hyperbolic space with curvature $-c^{0}\in \mathbb{R}$, where $c^{0}$ is learned in training.
Hyperbolic message passing
(Eqs.~\ref{equa:weight_multi} to \ref{equa:hyper_non_linear})
is then performed by multiple layers (forming the \emph{Hyperbolic Graph Embedding Layers} in Fig.~\ref{fig:model} (b)). The layer is indexed by $\ell$, ranging from $1$ to a pre-defined integer $l$.

\vspace{4pt}
\noindent (1) \textit{Hyperbolic Feature Transformation}
is performed by

\vspace{-5pt}
\begin{equation}
    \small
   \mathbf{m}_{i}^{\ell, H} = W^{\ell} \otimes_{c^{\ell-1}} \mathbf{x}_{i}^{\ell-1, H} \oplus_{c^{\ell-1}} \mathbf{b},
    \label{equa:weight_multi}
\end{equation}
where $W^{\ell} \in \mathbb{R}^{F^{\ell} \times F^{\ell-1}}$ is the 
weight
matrix,
and $\mathbf{b}\in \mathbb{D}_{c^{\ell-1}}^{F^{\ell}}$ denotes the bias (both are learned).
We employ a unique trainable curvature at \textit{each} layer to obtain a suitable hyperbolic space to account for different depths of the neural network.

\vspace{4pt}
\noindent (2) \textit{Hyperbolic Neighbor Aggregation.}  We then leverage the bridging between the hyperbolic space and the tangent space to perform neighbor aggregation~\cite{hyperbolicgat,gil}, resulting in $ \mathbf{h}_{i}^{\ell, H}\in \mathbb{D}_{c^{\ell-1}}^{F^{\ell}}$,

\vspace{-10pt}
\begin{equation}
    \small
    \mathbf{h}_{i}^{\ell, H}=\exp _{\mathbf{0}}^{c^{\ell-1}}\left(\sum_{j \in\{i\} \cup \mathcal{N}(i)} e_{i j} \log _{\mathbf{0}}^{c^{\ell-1}}\left(\mathbf{m}_{j}^{\ell, H}\right)\right).
    \label{equa:neigh_agg}
\end{equation}
$\mathcal{N}(i)=\{j:(i, j) \in \mathcal{E}\}$ denotes the set of neighbors of $i \in \mathcal{V}$. We apply out-degree normalization of $A$ (adjacency matrix), i.e.,  $D_{\text {out }}^{-1}(A+I)$, to obtain the aggregation weights
for simplicity
(while $e_{i j}$ can be computed with different mechanisms such as attention or leveraging edge attributes if present). $D_{\text {out }}$ is a diagonal matrix
such that element $(i,i)$
is the sum of row $i$ in $A$ plus $1$. We choose the tangent space of the origin for efficiency~\cite{hgcn}.

\vspace{4pt}
\noindent (3) \textit{Non-Linear Activation with Trainable Curvatures.}
$\mathbf{x}_{i}^{\ell, H}\in \mathbb{D}_{c^{\ell}}^{F^{\ell}}$, the output hyperbolic representation of node $i$ in layer $\ell$ is set as

\begin{equation}
    \small
    \mathbf{x}_{i}^{\ell, H}=\exp _{\mathbf{0}}^{c^{\ell}}\left(\sigma\left(\log _{\mathbf{0}}^{c^{\ell-1}}\left(\mathbf{h}_{i}^{\ell, H}\right)\right)\right).
    \label{equa:hyper_non_linear}
\end{equation}
To smoothly vary the curvature of each layer, in Eq.~\ref{equa:hyper_non_linear}, we first map $\mathbf{h}_{i}^{\ell, H}$ to the tangent space with the \logmap map. A point-wise 
non-linearity $\sigma(\cdot)$ (ReLU in our experiments) and an \expmap map are then used to bring the vector back to the hyperbolic space with a new learnable curvature $-c^{\ell}$.

\subsection{Neighborhood Collaborative Learning}
\label{sec:neighbor_disentangle}

Due to the semantics of directed edges, neighbors of a node can be implicitly partitioned into non-disjoint groups. Consideration of these neighborhoods is critical for learning a holistic node embedding in a digraph. This is because each neighborhood can reflect a distinct aspect pertaining to the node~\cite{disentangled}. E.g., in a social networking platform, a popular user's in-neighbors and out-neighbors can exhibit entirely different degrees of relationship cohesion to this popular user. Furthermore, users who share common in-neighbors with this user and those who share common out-neighbors can reveal additional contexts.  

Our method 
leverages this inductive bias exhibited in real-world digraphs through collaborative learning among the aforementioned four canonical types of neighborhoods in hyperbolic space. In addition, \ours achieves larger receptive fields by taking account of the impact of multi-ordered
neighbors~\cite{yang2017fast,dgcn}.
\ours generates a representation for each type and order of neighborhoods, each serving as one representation slice that eventually comprises the final holistic
node embedding by  collaboratively learning in the hyperbolic space.
With the neighbor-aggregation-based formalism (Eq.~\ref{equa:neigh_agg}) learning from multi-ordered
and partitioned neighborhoods, \ours is capable to model  general digraphs that contain cycles or non-transitive relations.

\noindent \textbf{Neighbor Partition.}
Four types of $k$-order proximity matrix are defined 
(Fig.~\ref{fig:model} (a)).
Formally, $k$-order proximity in terms of:

\noindent (1) \texttt{diffusion in} $A_{d_{in}}^{k}$,

\vspace{-10pt}
\begin{equation}
    \scriptsize
    A_{d_{in}}^{k} (i, j) =  \mathbb{1} \left(\sum_{p \in \mathcal{V}} A_{d_{in}}^{k-1} (i,p) \cdot A_{d_{in}}^{1}(p,j)   \right)
\end{equation}
where $A_{d_{in}}^{1} = A^\intercal$, $\cdot$ is the inner product and $\mathbb{1}$ is the indicator function. $A_{d_{in}}^{k} (i, j)=1$ if there is a directed path from node $j$ to node $i$ of length exactly $k$.

\noindent (2) \texttt{diffusion out} $A_{d_{out}}^{k}$,

\vspace{-5pt}
\begin{equation}
    \scriptsize
    A_{d_{out}}^{k} (i, j) =  \mathbb{1} \left(\sum_{p \in \mathcal{V}} A_{d_{out}}^{k-1}(i, p)  \cdot A_{d_{out}}^{1} (p, j)  \right)
\end{equation}
where $A_{d_{out}}^{1}=A$. $A_{d_{out}}^{k} (i, j)=1$ if there is a directed path from node $i$ to node $j$ of length exactly $k$.

\noindent (3) \texttt{common in} $A_{c_{in}}^{k}$,

\vspace{-15pt}
\begin{equation}
    \scriptsize
    A_{c_{in}}^{k} (i, j) =  \mathbb{1} \left(\sum_{p \in \mathcal{V}} A_{d_{in}}^{k}(i,p) \cdot A_{d_{out}}^{k}(p,j) \right)
\end{equation}
where $i\neq j\neq p$. $A_{c_{in}}^{k} (i, j)=1$ if  node $i$ and node $j$ have a common in-neighbor $k$ hops away.

\noindent (4) \texttt{common out} $A_{c_{out}}^{k}$,

\vspace{-5pt}
\begin{equation}
    \scriptsize
    A_{c_{out}}^{k} (i, j) =  \mathbb{1} \left(\sum_{p \in \mathcal{V}} A_{d_{out}}^{k}(i,p) \cdot A_{d_{in}}^{k}(p,j) \right)
\end{equation}
where $i\neq j\neq p$. $A_{c_{out}}^{k} (i, j)=1$ if node $i$ and node $j$ have a common out-neighbor $k$ hops away. 

\vspace{4pt}
\noindent \textit{Multi-Scale Neighborhood Learning}. For a given non-zero integer $K$, we compute the four types of $k$-order proximity matrix for $k=1$ to $K$ (as a preprocessing step). 
This enables capturing multi-scale node proximity and the nodes' local directed graph characteristics.
These $k$-order proximity matrices replace the original adjacency matrix $A$ to provide a wider range of neighborhoods to Hyperbolic Graph Embedding Layers
(Fig.~\ref{fig:model} (b)).

\noindent \textbf{Neighborhood Aggregation.} We then apply \textit{Hyperbolic Neighborhood Aggregation}
to enable 
a joint assessment of
the  neighborhoods.
Here, we view the $4K$ output hyperbolic vectors from the Hyperbolic Graph Embedding Layers as representations of $4K$ \emph{neighbors} of the anchor node $i$. We consider $\mathbf{z}_i^{\text{fuse}}$, which is the hyperbolic average of these $4K$ vectors, as the initial representation of node $i$ before hyperbolic neighborhood collaboration.
Subsequently, we apply Eq.~(\ref{equa:neigh_agg})
with the learned curvature $-c^{l}$ 
from the last hyperbolic graph embedding layer $l$, and use equal aggregation weights $ \frac{1}{4K+1}$ ($4K$ plus $1$ because $\mathbf{z}_i^{\text{fuse}}$ itself is included as a neighbor in order to enforce a skip connection).
The resulting output, $\mathbf{z}_i^{l,H}$, is the final hyperbolic embedding of node $i$.
Hyperbolic Neighborhood Aggregation can encourage a better utilization of neighborhoods by synthesizing intermediate representations learned in a neighborhood-level in hyperbolic space.

\subsection{Self-Supervised Learning with Asymmetry-Preserving Regularizers }  
\label{sec:asymmetry}

\textit{Homophily} and \textit{preferential attachment} are two driving forces of link formation according to sociopsychological theories. Homophily~\cite{birds} refers to the notable role of similarity, 
often summarized as ``birds of a feather flock together'', and preferential attachment~\cite{mitzenmacher2004brief} describes the role of prior connectivity: the link formation likelihood is asymmetric and determined by individual connectivity.
To model these two decomposed causes of link formation,
we invoke two
regularizers to predict directed edges when training \ourseos, thus allow it to respect asymmetry in digraph link formation
by learning it as a self-supervised task.

We first adopt the Fermi-Dirac decoder~\cite{krioukov2010hyperbolic} as a regularizer to reinforce the learning of an appropriate node-pair distance in the hyperbolic embedding space (to well account for homophily). The hyperbolic Fermi-Dirac decoder defines the likelihood of a node pair $(i, j)$ as 

\vspace{-20pt}
\begin{equation}
    \scriptsize
    p(i, j)_f = \frac{1}{e^{\left(d_{\mathbb{D}_{c^{l}}^{d'}}\left(\mathbf{z}_{i}^{l, H}, \mathbf{z}_{j}^{l, H}\right)^{2}-r\right) / t}+1},
    \label{equa:fermi}
\end{equation}
where $r$=$2$ and $t$=$1$ (default),
and $d_{\mathbb{D}_{c^{l}}^{d'}}(\cdot, \cdot)$ is the hyperbolic distance (Eq.~\ref{equa:dist}).

We further preserve the individual asymmetric node connectivity by learning an additional $1$-dimensional mass for each node.
This design is elegantly derived from Newton’s theory of universal gravitation:
each particle in the universe attracts other particles through gravity, which is proportional to 
their masses, and inversely proportional to their distance.
The learnable node mass is flexible, and it encompasses many centrality measures, including Katz, Betweenness and Pagerank. 
It is also capable of providing explainable visualizations~\cite{gravity}. 
To incorporate this idea into \ours based in hyperbolic space instead of Euclidean, 
we map $\mathbf{z}_{i}^{l, H}$
to the tangent space of the origin with the logarithmic map
(i.e., $\mathbf{z}_{i}^{l, E} = \log _{\mathbf{0}}^{c^{l}}\left(\mathbf{z}_{i}^{l, H}\right)$), and then employ a Euclidean linear layer to learn $m_i\in\mathbb{R}$ (mass of node $i$). The likelihood of node pair $(i, j)$ is
computed by

\vspace{-10pt}
\begin{equation}
\scriptsize
    p(i, j)_g = \gamma\left(m_{j}-\lambda \log \left(d_{\mathbb{D}_{c^{l}}^{d'}}(\mathbf{z}_{i}^{l, H}, \mathbf{z}_{j}^{l, H})^2 \right)\right),
    \label{equa:gravity}
\vspace{-3pt}
\end{equation}
where $\gamma$ denotes the sigmoid function, and $\lambda\in \mathbb{R}$ is a hyper-parameter that weights the relative importance of the symmetric embedding distance to the asymmetric node relationships. 
$p(i, j)_g\neq p(j, i)_g$.
Eqs.~(\ref{equa:gravity}) and (\ref{equa:fermi}) both serve as self-supervised regularizers by minimizing the binary cross-entropy loss with negative sampling to estimate the likelihood of each node pair. 
However, the two are placed at different depths of \ourseos. Specifically, 
Eq.~(\ref{equa:gravity}) is employed one layer after where Eq.~(\ref{equa:fermi}) is used. 
Thus, even though $d_{\mathbb{D}_{c^{l}}^{d'}}(\cdot, \cdot)$ also appears in Eq.~(\ref{equa:gravity}),  we 
find that Eq.~(\ref{equa:fermi}) provides auxiliary guidance for the model to better construct the final hyperbolic embedding space.

\subsection{Time Complexity}
The time complexity of the the \textit{Hyperbolic Graph Embedding Layer} is  \small{$O(K   n   d^{\ell-1}   d^{\ell}+ K   m   d^{\ell})$} \normalsize
where $K$ denotes the maximal order of the $k$-order proximity matrix. $d^{\ell-1}$ and  $d^{\ell}$, respectively, denote the dimensionality of input and output features of layer $\ell$. $n$ and $m$ are the number of nodes and edges respectively. The time complexity of  \textit{Hyperbolic Neighborhood Aggregation} is   \small{$O(K   n   d^{l}   d^{l})$}\normalsize, where $d^{l}$ denotes the dimensionality of output features of the final layer $l$.
Supposing $d^{\ell-1}$ and $ d^{\ell}$ are equal to $d$, an $l$-layer model has a cost of \small{$O(lKnd^2+lKmd)$}\normalsize. The time complexity is on par with other GNN methods such as HAT and GCN that have a complexity of \small{$O(lnd^2+lmd)$}\normalsize, because in practice $K$ is a small non-negative integer (e.g., the maximum $K$ is $3$ in our paper, and most of the time, setting $K$ to $2$ would be sufficient).

\section{Experimental Setups}
\label{sec:exp_setup}

\noindent \textbf{Datasets. }
We use open access homogeneous \emph{digraph} datasets of varied size and 
domain
(Table~\ref{tab:datasets}),
and create numerous splits
of each dataset and task for more reliable results.

\noindent \textbf{Tasks \& Metrics. }
We use the following tasks and metrics.

\vspace{-3pt}
\begin{itemize}[leftmargin=*]
    \item \textit{Link Prediction (LP).} LP  demonstrates 
    a method's  capability
    in 
    modeling 
    asymmetric node connectivity, as a binary classification task of discriminating 
    the missing edges from  
    the fake ones. Given a digraph $\mathcal{G}$, we train models on its incomplete version $\mathcal{G}^\prime$ by randomly removing edges. Half of the removed edges form the positive samples in the validation set, and the other half form the positive samples in the test set.  
The negative samples are randomly sampled from unconnected node pairs in $\mathcal{G}$, drawing the same number as there are positive samples.
    Metrics are 
    AUC (Area under the ROC Curve) and AP (Average Precision).

    \item  \textit{Semi-supervised Node Classification (NC)~\cite{digcn}.} In NC, each dataset contains only $20$ labeled nodes for each node class, which requires use of the graph structure for predicting the labels of remaining nodes. The validation set consists of $500$ random unlabeled nodes. Unlabeled nodes not in the validation set make up the test set.
    
    \item \textit{Link Sign (Property) Prediction (SP).} Many real-world graphs are \textit{signed networks}, e.g., social networks that allow trust and distrust user relationships.
    We use the Wiki dataset to evaluate the accuracy in
    predicting attributes of directed edges representing votes $\{\text{oppose}, \text{neutral}, \text{support}\}$~\cite{side}.
    Given a digraph $\mathcal{G}$, $5\%$ of  edges are labeled for training, $5\%$ for validation, and $90\%$ for testing.

    \item  \textit{Embedding Visualization (EV).} EV
    shows
    the expressiveness of methods qualitatively. We visualize node representations in 
    $2$D space projected via PCA~\cite{pca}.
    Embedding vectors are obtained from the  NC task.
    Hyperbolic embeddings are mapped to the Euclidean space before $2$D projection.
\end{itemize}

\noindent In all tables, the best score is \red{\textbf{bolded}}, the second best is \blue{\ul{underlined}}, and the third best is in \magenta{\textit{italic}}. Relative gains are computed as $(\textsc{Best}-\textsc{Second})/\textsc{Second}$. 
$^\ast$ indicates statistically superior performance of the best to the second best
at a significance level of $0.001$ using a standard paired t-test. Values after $\pm$ are standard deviations.

\begin{table}[t]
\begin{center}
\caption{Statistics of datasets. Reciprocity 
measures
the likelihood of nodes
to be mutually linked.
Label rate is the ratio
of nodes labeled for training.
}
\vspace{-8pt}
\label{tab:datasets}
\setlength{\tabcolsep}{1.8pt}
\fontsize{7.6pt}{7.6pt}\selectfont
\begin{tabular}{lrrrrrrr}
\toprule
\multirow{2}{*}{\textbf{LP}} & \multirow{2}{*}{\textbf{Reciprocity}} & \multirow{2}{*}{\textbf{\# Nodes}} & \multirow{2}{*}{\textbf{\# Edges}} & \multirow{2}{*}{\textbf{Nodes}} & \multirow{2}{*}{\textbf{Edge}}  & \multicolumn{2}{c}{\textbf{Degree}}  \\  
    &    &  &  &  &     & \footnotesize{Avg.}  & \footnotesize{Max}  \\ \midrule
Air & $15.68\%$ & $1,226$ & $2,615$ & Airport & Preferred Route & $4$ & $37$  \\
Blog & $24.25\%$ & $1,224$ & $19,025$ & Blog & Hyperlink & $31$ & $467$   \\ 
Survey & $38.77\%$ & $2,539$ & $12,969$ & User &  Friendship & $10$ & $36$  \\
Cora  & $0.06\%$ & $2,708$   & $5,429$ & Paper & Citation & $4$ & $168$    \\
DBLP  & $0.43\%$ & $12,591$  & $49,743$ & Paper & Citation  & $8$ & $710$   \\
\bottomrule
\end{tabular}
\begin{tabular}{lrrrrrrr}
\multirow{2}{*}{\textbf{NC}}  & \multirow{2}{*}{\textbf{\# Nodes}} & \multirow{2}{*}{\textbf{\# Edges}} & \multirow{2}{*}{\textbf{Node}} & \multirow{2}{*}{\textbf{Edge}}  & \multirow{2}{*}{\textbf{Classes}}  & \multirow{2}{*}{\textbf{Features}}   & \multirow{2}{*}{\textbf{Label Rate}}   \\ 
    &    &  &   &  &    &  & \\ \midrule
CiteSeer & $3,312$ & $4,715$ & Paper & Citation  & $6$ & $3,703$ & $3.62\%$   \\
Cora-ML & $2,995$ & $8,416$ & Paper & Citation & $7$ & $2,879$ & $4.67\%$    \\
Wiki  & $7,115$   & $103,689$ & User & Vote  & $3$  & $7,115$   &  $0.84\%$   \\
\bottomrule
\end{tabular}
\end{center}
\vspace{-8pt}
\end{table}

\begin{table*}[t]
\begin{center}
\caption{Results of \textit{Link Prediction on Digraphs} with 4- or 8-dimensional node embeddings.
} 
\vspace{-10pt}
\label{tab:lp_result_fewdim}
\setlength{\tabcolsep}{2.5pt}
\fontsize{7.5pt}{7.5pt}\selectfont
\begin{tabular}{@{}lcccccccc@{}}
\toprule
\multirow{3}{*}{\textbf{Model (4/8-Dim)}}   & \multicolumn{4}{c}{\textbf{Air}} &  \multicolumn{4}{c}{\textbf{Cora}}  \\ \cline{2-9}
  &  \multicolumn{2}{c}{\textbf{4-Dim}}   &  \multicolumn{2}{c}{\textbf{8-Dim}}   &  \multicolumn{2}{c}{\textbf{4-Dim}}   &  \multicolumn{2}{c}{\textbf{8-Dim}} \\ 
  & AUC   & AP   & AUC   & AP  & AUC   & AP  & AUC  & AP \\ \midrule
GCN~\cite{gcn}   & 67.88 ($61.73$)  & 67.88 ($60.51$)  & 69.21 ($64.05$)  & 69.68  ($63.48$) &  65.92 ($61.00$) & 65.92 ($59.97$)   & 70.89  ($65.67$) & 71.26  ($65.28$) \\
VGAE~\cite{vgae}   & 69.77 ($62.86$)  & 70.73 ($62.55$) &  73.49 ($66.87$)  &    74.04 ($66.95$) &  63.86 ($56.90$) & 63.86 ($55.39$)  &  66.60 ($60.33$) & 66.60 ($58.75$)  \\
GAT~\cite{gat}    & 69.02 ($63.48$)  &  69.02 ($62.86$)  & 71.31 ($67.03$)  &  71.31 ($67.01$)  &  68.18 ($64.73$) &  68.18 $64.31$)   &  72.70 ($68.70$) &  73.93 ($69.08$)  \\
Gravity GCN $^\text{\textdagger}$~\cite{gravity}     & 65.20 ($59.41$)  &  67.73 ($60.98$)  &  74.00  ($68.91$)   & 75.43 ($69.14$)  & 70.37 ($65.80$)  & 70.37 ($64.65$)  & 75.29 ($71.85$)  &  77.17 ($72.50$)  \\
Gravity VGAE $^\text{\textdagger}$~\cite{gravity}   & 62.24 ($55.48$)  & 62.24 ($54.97$)  &  68.00 ($60.23$)   & 68.00 ($59.57$)  & 66.74 ($61.79$) & 66.74 ($60.61$)  & 71.04 ($65.45$) & 71.04 ($64.15$)  \\
DGCN $^\text{\textdagger}$~\cite{dgcn}    & 74.36  ($65.75$)  &  71.42   ($63.27$) &  77.23 ($70.60$) & 75.86 ($70.27$)   & 75.33  (\magenta{\textit{71.88}})   &  71.95  (\magenta{\textit{68.58}}) &  79.01  ($75.30$) & 79.01 ($74.28$)   \\ 
DiGCN $^\text{\textdagger}$~\cite{digcn}     & 72.59 ($64.37$)  & 70.01  ($61.66$)    &  74.65 ($69.27$)  &  75.40 ($68.29$) & 70.61  ($65.81$)   &  67.11 ($61.57$)  & 74.63 ($70.65$)  &  74.88 ($69.86$) \\ 
MagNet $^\text{\textdagger}$~\cite{zhang2021magnet}     &  72.26  ($58.44$)   &   71.10 ($57.92$) &  76.64  ($64.26$)  & 78.62   ($64.69$)  & \magenta{\textit{77.45}} ($55.93$) & \magenta{\textit{79.32}} ($56.84$)   & 77.46 ($66.82$)   & 76.59 ($63.96$) \\ 
HAT $^\text{\textsection}$ ~\cite{hyperbolicgat}    & \magenta{\textit{76.11}} (\blue{\ul{71.24}}) & \magenta{\textit{73.72}} (\blue{\ul{69.35}})  &   \magenta{\textit{80.52}} (\magenta{\textit{75.13}})  & \magenta{\textit{79.73}} (\magenta{\textit{74.05}})  & 76.25 (\blue{\ul{72.84}})   & 74.38 (\blue{\ul{70.27}})    &  \magenta{\textit{82.58}} (\magenta{\textit{77.82}})  &  \magenta{\textit{82.05}} (\magenta{\textit{77.39}})  \\  
HGCN $^\text{\textsection}$ ~\cite{hgcn}    & \blue{\ul{80.90}} (\magenta{\textit{66.63}})  &  \blue{\ul{80.90}} (\magenta{\textit{65.95}}) & \blue{\ul{84.67}} (\blue{\ul{77.65}})   & \blue{\ul{85.97}} (\blue{\ul{78.14}})   &  \blue{\ul{80.02}} ($67.37$) & \blue{\ul{82.16}} ($66.66$) &  \blue{\ul{85.05}} (\blue{\ul{83.07}})    &  \blue{\ul{88.04}} (\blue{\ul{84.63}})  \\ \midrule
\ours (ours) $^\text{\textdagger}$$^\text{\textsection}$      & \textbf{\red{85.79} ($^\ast$\red{81.69})} & \textbf{\red{85.92} ($^\ast$\red{81.93})}  &  \textbf{\red{88.46} ($^\ast$\red{84.26})}  &  \textbf{\red{88.46}  ($^\ast$\red{84.82})}   &  \textbf{\red{86.08} ($^\ast$\red{83.99})} & \textbf{\red{88.74} ($^\ast$\red{85.33})} & \textbf{\red{88.88} ($^\ast$\red{86.31})}  & \textbf{\red{91.13} ($^\ast$\red{87.76})}  \\ \midrule
Relative Gains ($\%$)    &  6.04 (14.67)  &  6.21 (18.14) & 4.48 (8.51)   &  2.90 (8.55)  & 7.57 (15.31)   & 8.01 (21.43)    & 4.5 (3.9)  & 3.51 (3.7)  \\ \bottomrule
\end{tabular}
\end{center}
\begin{flushleft}\small{Note: $^\text{\textdagger}$ denotes the method was designed specifically for homogeneous digraphs (i.e., DRL), and $^\text{\textsection}$  denotes the use of hyperbolic space. Results (in percentage $\%$) on each dataset of each method are from $100$ repeated experiments ($10$ different train/test splits per dataset and $10$ runs using different random seeds per split). We list the best and the average results, and the average is shown in  brackets. 
}\end{flushleft} 
\end{table*}

\begin{table*}[t]
\begin{center}
\caption{Results of \textit{Link Prediction on Digraphs} with 32-dimensional node embeddings. 
} 
\vspace{-10pt}
\label{tab:lp_result}
\setlength{\tabcolsep}{1.5pt}
\fontsize{7.1pt}{7.1pt}\selectfont
\begin{tabular}{@{}lcccccccccc@{}}
\toprule
\multirow{2}{*}{\textbf{Model (32-Dim)}}   & \multicolumn{2}{c}{\textbf{Air}}  & \multicolumn{2}{c}{\textbf{Cora}} & \multicolumn{2}{c}{\textbf{Blog}} & \multicolumn{2}{c}{\textbf{Survey}} & \multicolumn{2}{c}{\textbf{DBLP}}  \\ \cline{2-11}
  & AUC   & AP   & AUC   & AP  & AUC   & AP  & AUC  & AP & AUC  & AP   \\ \midrule
MLP  & 81.29   ($76.52$)  & 83.53 ($78.18$) & 84.47 ($81.67$) & 87.70 ($83.69$)  & 93.31 ($92.48$)  & 93.31 ($92.45$)  & 91.21 ($89.98$) & 92.46 ($90.75$)  &  51.22 ($49.98$) & 51.22 ($49.99$) \\
NERD $^\text{\textdagger}$~\cite{nerd}  & 60.62 ($56.39$) & 67.37 ($60.19$)  & 65.62 ($62.02$) & 71.68 ($65.66$) &  95.03 ($94.00$) &   95.03 ($93.47$) & 77.12 ($69.30$) & 79.60 ($70.80$)  & 95.78  ($95.37$) & 95.93 ($95.41$)\\
ATP $^\text{\textdagger}$~\cite{atp} & 68.99 ($66.40$) &  68.99 ($64.99$)   &  88.47 (\magenta{\textit{86.44}})  & 88.47 ($86.04$) & 85.05 ($83.46$)  & 85.05 ($79.30$) & 73.53  ($71.47$)   &  73.53 ($70.64$)  & 60.43  ($59.21$)  & 60.43  ($57.37$) \\
APP $^\text{\textdagger}$~\cite{app} & 85.08 ($82.72$) & 86.35 ($84.58$)    &  86.65 ($85.50$) & 89.80 ($87.22$)  & 92.33  ($91.65$) & 92.33 ($90.55$) & 91.16 ($90.34$)  & 92.77 ($91.14$)  & 95.58 ($95.33$) & 9573 ($95.41$) \\
GCN~\cite{gcn}   &  76.71 ($72.27$) & 80.95 ($75.13$) & 80.77 ($78.73$) & 85.67 ($81.21$)  & 91.87  ($90.18$) & 92.16 ($90.54$) & 89.29 ($87.98$)  & 91.78 ($89.42$)  & 92.98 ($92.34$) & 94.37 ($93.15$)  \\
VGAE~\cite{vgae}   &  77.79 ($73.75$)  & 82.73 ($76.75$) & 80.80 ($79.24$) & 85.47 ($81.57$)  & 92.25 ($91.39$)  &  92.80 ($91.85$) & 90.07 ($88.78$)  & 92.39 ($90.14$)   & 93.36 ($92.64$)  & 94.85 ($93.45$) \\
GAT~\cite{gat}    & 84.21 ($80.24$) & 84.79 ($81.46$)  & 85.40 ($82.58$) & 88.53 ($84.60$)  &  92.69 ($89.95$) & 92.69 ($89.83$) & 92.01 ($91.05$ ) & 93.09 ($91.65$)   & 95.94 ($95.62$) & 96.28 ($95.80$)  \\
Gravity GCN $^\text{\textdagger}$~\cite{gravity}      & 85.16 ($82.22$) & 86.86 ($83.50$) & 85.62 ($83.87$) & 88.73  ($85.62$)  &  95.11 ($94.46$) & 95.11 ($94.31$) & 91.63 ($90.86$) & 93.11 ($91.76$)  & 96.89  ($96.78$) & \magenta{\textit{97.46}}  (\magenta{\textit{97.34}}) \\
Gravity VGAE $^\text{\textdagger}$~\cite{gravity}  & 83.98 ($80.06$) &  85.67 ($81.61$)   & 87.17 ($84.46$) & 89.51 ($86.22$) & \blue{\ul{96.15}} (\blue{\ul{95.59}})  & \blue{\ul{96.15}} (\blue{\ul{95.42}}) & 91.64 ($90.96$) & 93.23 ($91.82$)   & 95.98 ($95.57$)  & 96.24 ($95.81$) \\
DGCN $^\text{\textdagger}$~\cite{dgcn}  & 77.83 ($73.68$) & 80.79 ($75.64$)   & 83.57 ($81.34$) & 85.48 ($83.00$)  & 87.74 ($86.74$)  &  88.13 ($86.75$)  & 90.47 ($89.49$)  &  91.27 ($89.94$)   & 92.26 ($91.83$) & 90.16 ($89.52$) \\
DiGCN$^\text{\textdagger}$~\cite{digcn}   & 75.35 ($71.27$)  & 77.64 ($73.97$)    &  81.80 ($78.90$) & 83.03 ($79.92$)  &  91.98 ($90.50$)  & 89.34 ($87.36$)  & 89.85 ($88.17$) & 89.80 ($88.08$)   & 89.99 ($89.72$) &  89.93 ($89.60$)  \\
MagNet $^\text{\textdagger}$~\cite{zhang2021magnet}  & 79.32 ($75.58$)  & 80.66 ($76.34$)   & 82.77 ($71.90$) &    81.63 ($69.84$)  &  91.83 ($90.81$) &  90.46 ($89.29$)  & 86.65  ($84.81$) & 87.76 ($85.71$)  & 81.89 ($80.57$)  & 81.68 ($81.50$) \\
HNN $^\text{\textsection}$~\cite{hnn}  & \blue{\ul{88.42}} (\magenta{\textit{85.79}})  & \blue{\ul{88.95}} (\magenta{\textit{86.40}})    & \magenta{\textit{88.75}} ($86.33$)  & \magenta{\textit{90.81}} (\magenta{\textit{87.81}}) & \magenta{\textit{95.80}} (\magenta{\textit{95.39}}) & \magenta{\textit{95.80}} (\magenta{\textit{95.16}}) & \magenta{\textit{92.07}} (\magenta{\textit{91.39}})  & \blue{\ul{93.40}} (\magenta{\textit{92.04}}) & \magenta{\textit{97.43}} (\magenta{\textit{97.14}}) & 97.43 ($97.13$)\\
HGCN $^\text{\textsection}$~\cite{hgcn}  & \magenta{\textit{88.26}} (\blue{\ul{86.12}})  & \magenta{\textit{88.88}} (\blue{\ul{86.64}})  & \blue{\ul{89.24}} (\blue{\ul{87.68}})  & \blue{\ul{91.54}} (\blue{\ul{88.97}} ) & 95.64 ($95.23$)  & 95.64 ($95.00$) & \blue{\ul{92.15}} (\blue{\ul{91.50}}) & \magenta{\textit{93.38}} (\blue{\ul{92.08}}) & \blue{\ul{97.54}} (\blue{\ul{97.33}}) &  \blue{\ul{97.62}}  (\blue{\ul{97.37}}) \\ \midrule
\ours (ours) $^\text{\textdagger}$$^\text{\textsection}$    &  \red{\textbf{89.07}} (\red{\textbf{86.33}}) &  \red{\textbf{89.21}} ($^\ast$\red{\textbf{86.86}})    & \red{\textbf{89.50}} ($^\ast$\red{\textbf{88.22}})  & \red{\textbf{91.62}} ($^\ast$\red{\textbf{89.47}}). & \red{\textbf{96.19}} (\red{\textbf{95.62}})  & \red{\textbf{96.18}} ($^\ast$\red{\textbf{95.48}})  & \red{\textbf{92.56}} ($^\ast$\red{\textbf{91.96}}) & \red{\textbf{93.63}} ($^\ast$\red{\textbf{92.46}})   &  \red{\textbf{97.66}} ($^\ast$\red{\textbf{97.38}})   &  \red{\textbf{97.75}} ($^\ast$\red{\textbf{97.44}})  \\ \midrule
Relative Gains ($\%$)    & 0.74 (0.24)  & 0.29 (0.25)   &  0.29 (0.62)     &  0.09 (0.56) & 0.04 (0.03) &  0.03 (0.06)  & 0.44 (0.50)  & 0.25 (0.41)    &  0.12 (0.05)  & 0.13  (0.07)  \\ \bottomrule
\end{tabular}
\end{center}
\begin{flushleft}\small{Note: Every result is from $100$ experiments (the same as in Table~\ref{tab:lp_result_fewdim}). }\end{flushleft} 
\end{table*}

\begin{table}[t]
\vspace{-5pt}
\begin{center}
\caption{Results of \textit{Node Classification on Digraphs}.
\label{tab:nc_result}
} 
\vspace{-10pt}
\label{tab:nc_result}
\fontsize{7.5pt}{7.5pt}\selectfont
\begin{tabular}{@{}llrr@{}}
\toprule
   &  \textbf{Model} & \textbf{CiteSeer}  &  \textbf{Cora-ML}\\ \midrule
  \multirow{7}{*}{4-Dim} &  MLP     & $37.68\pm3.0$     &  $51.19\pm6.3$  \\
&  GCN~\cite{gcn}           &  $32.82\pm7.9$    &  \textit{60.56$\pm$9.8} \\
&  GAT~\cite{gat}         & \magenta{\textit{51.97$\pm$4.2}}   & \magenta{\textit{68.38$\pm$3.4}}\\
&  DGCN~\cite{dgcn}         & $38.67\pm10.0$  & $53.44\pm11.1$  \\ 
&  DiGCN~\cite{digcn}       &   \blue{\ul{53.43$\pm$10.3}}   & \blue{\ul{71.35$\pm$2.3}} \\ 
&  HNN~\cite{hnn}       &  $47.44\pm2.9$  &  $52.76\pm4.9$ \\
&  HGCN~\cite{hgcn}      & $42.24\pm3.6$   &    $52.17\pm5.9$ \\ \cline{2-4}
&  \ours (ours)           &  \red{\textbf{$^\ast$65.72$\pm$2.9}}   &   \red{\textbf{$^\ast$74.63$\pm$1.2}}  \\ \cline{2-4}
&  Relative Gains ($\%$)      &  23.00     &  4.60 \\ \midrule
  \multirow{7}{*}{8-Dim} &  MLP    &  $51.70\pm2.6$     &  $60.48\pm1.8$ \\
&  GCN~\cite{gcn}         & $36.26\pm6.5$   &  $67.62\pm10.8$  \\
&  GAT~\cite{gat}          & $50.81\pm3.9$   &  $74.87\pm1.8$ \\
&  DGCN~\cite{dgcn}         & \magenta{\textit{57.27$\pm$2.4}}  & \magenta{\textit{77.16$\pm$4.4}}  \\ 
&  DiGCN~\cite{digcn}       &  \blue{\ul{60.37$\pm$2.6}}   & \blue{\ul{78.38$\pm$1.2}} \\ 
&  HNN~\cite{hnn}      &    $50.73\pm3.1$  &   $61.54\pm2.1$  \\
&  HGCN~\cite{hgcn}       &  $52.57\pm2.3$   &  $73.44\pm2.3$ \\ \cline{2-4} 
&  \ours (ours)        &    \red{\textbf{$^\ast$67.96$\pm$1.6}}      &    \red{\textbf{$^\ast$81.55$\pm$1.6}} \\ \cline{2-4}
&  Relative Gains ($\%$)      & 12.57   &  4.04 \\ \midrule
  \multirow{7}{*}{32-Dim} &  MLP     & $53.18\pm 1.6$  & $61.63\pm 1.8$  \\
&  GCN~\cite{gcn}  & $53.20\pm 3.1$    & $69.51\pm 8.5$     \\
&  GAT~\cite{gat}       & $63.03\pm 0.6$     & $71.91\pm0.9$\\
&  DGCN~\cite{dgcn}       & \magenta{\textit{64.17$\pm$2.4}}    & \blue{\ul{81.29$\pm$1.6}}  \\ 
&  DiGCN~\cite{digcn}       & \blue{\ul{65.83$\pm$1.8}}    & \magenta{\textit{78.08$\pm$1.9}} \\
&  HNN~\cite{hnn}     & $56.10\pm 2.2$    &   $62.49\pm 2.6$  \\
&  HGCN~\cite{hgcn}     &  $59.02\pm 2.3$    &  $76.48\pm1.5$ \\ \cline{2-4}
&  \ours (ours)      &     \red{\textbf{$^\ast$70.66 $\pm$ 1.2}}   &   \red{\textbf{$^\ast$82.19 $\pm$ 1.3} }   \\ \cline{2-4}
&  Relative Gains ($\%$)      & 7.34  &   1.11 \\ \midrule
  \multirow{7}{*}{64-Dim} &  MLP    &     $57.20\pm1.9$  & $65.43\pm2.9$ \\
&  GCN~\cite{gcn}          &  $52.71\pm4.1$     &  $72.53\pm2.0$\\
&  GAT~\cite{gat}         &  $56.29\pm2.5$     & $75.50\pm1.5$ \\
&  DGCN~\cite{dgcn}       & \blue{\ul{64.45$\pm$1.6}}   &  \blue{\ul{80.93$\pm$1.8}}  \\ 
&  DiGCN~\cite{digcn}     &  \magenta{\textit{62.88$\pm$7.5}}     &  \magenta{\textit{79.90$\pm$1.1}}  \\
&  HNN~\cite{hnn}    &   $55.80\pm1.9$    &  $65.82\pm2.2$ \\
&  HGCN~\cite{hgcn}   &   $58.73\pm2.8$   &  $76.49\pm1.3$  \\ \cline{2-4} 
&  \ours (ours)        &  \red{\textbf{$^\ast$69.07$\pm$1.5}}   &  \red{\textbf{$^\ast$81.20$\pm$1.1}} \\ \cline{2-4}  
&  Relative Gains ($\%$)      &   7.17  &    0.33  \\  \midrule
  \multirow{7}{*}{128-Dim} &  MLP    &  $57.68\pm1.8$      &  $66.29\pm2.2$\\
&  GCN~\cite{gcn}         & \magenta{\textit{57.87$\pm$2.4}}      & $73.84\pm2.4$ \\
&  GAT~\cite{gat}      &  $56.48\pm2.1$        & $74.82\pm1.8$\\
&  DGCN~\cite{dgcn}     & \blue{\ul{66.25$\pm$1.5}}       &  \blue{\ul{81.50$\pm$1.6}} \\ 
&  DiGCN~\cite{digcn}        &   $56.50\pm14.1$    & \magenta{\textit{79.83$\pm$1.2}}\\ 
&  HNN~\cite{hnn}    &    $56.23\pm2.4$   & $65.12\pm1.7$ \\
&  HGCN~\cite{hgcn}    &   $57.65\pm3.2$    & $76.92\pm 1.6$ \\ \cline{2-4} 
&  \ours (ours)        &  \red{\textbf{$^\ast$70.53$\pm$1.1}}   &  \red{\textbf{$^\ast$81.77$\pm$1.3}}  \\ \cline{2-4}
&  Relative Gains ($\%$)      &  6.46  & 0.33   \\ \midrule
  \multirow{7}{*}{256-Dim} & MLP      &  $57.26\pm2.2$  &  $64.86\pm3.1$ \\
&  GCN~\cite{gcn}          & $55.82\pm3.2$    & $75.20\pm1.9$  \\
&  GAT~\cite{gat}      &  $57.66\pm2.4$         & $74.19\pm1.5$\\
&  DGCN~\cite{dgcn}       & \blue{\ul{65.90$\pm$1.5}}    &  \blue{\ul{81.29$\pm$ 1.4}}  \\ 
&  DiGCN~\cite{digcn}      & $46.36\pm 13.75$      & \magenta{\textit{79.46$\pm$1.2}}\\  
&  HNN~\cite{hnn}      &  $54.64\pm2.4$ & $66.09\pm2.0$  \\
&  HGCN~\cite{hgcn}     &  \magenta{\textit{58.23$\pm$ 2.3}}   &  $76.91\pm 1.7$\\ \cline{2-4} 
&  \ours (ours)         &  \red{\textbf{$^\ast$71.10$\pm$1.2}}  &  \red{\textbf{$^\ast$81.80$\pm$ 1.4}}  \\ \cline{2-4}  
&  Relative Gains ($\%$)      &  7.89  &  0.63  \\ \midrule \toprule
  \multirow{6}{*}{Results in~\cite{dgcn} (32-Dim) } &  ChebNet~\cite{chebnet}    & $56.46\pm 1.4$    & $64.02\pm1.5$ \\
&  SGC~\cite{sgc}         & $44.07\pm 3.5$  & $51.14\pm 0.6$ \\
&  APPNP~\cite{appnp}        & $65.39\pm 0.9$   & $70.07\pm 1.1$\\
&  InfoMax~\cite{infomax}   & $60.51\pm 1.7$      & $58.00\pm 2.4$\\
&  GraphSage~\cite{graphsage}    & $63.19\pm 0.7$   & $72.06\pm 0.9$\\
&  SIGN~\cite{sign}        & $60.69\pm 0.4$ & $66.47\pm0.9$ s\\   \bottomrule
\end{tabular}
\end{center}
\begin{flushleft}\small{Note: $20$ random splits per dataset are used for this task.}\end{flushleft} 
\vspace{-5pt}
\end{table}

\noindent \textbf{Implementation Details.}
Hyperparameter tuning was performed for each method per task and dataset (on the first split), which \textit{substantially} improved the results of ablations and baselines.
We searched initial learning rates \small{$\{0.001, 0.01, 0.1\}$}\normalsize, momentums \small{$\{0.9, 0.999\}$}\normalsize, weight decays \small{$\{0, 0.001\}$}\normalsize, and dropout rates \small{$\{0, 0.05, 0.1\}$}\normalsize.
Unique hyperparameters associated with each method were considered as well. E.g., for GAT, we searched the number of attention heads from \small{$\{4, 8\}$}\normalsize~and $\alpha$ from \small{$\{0.1, 0.2\}$}\normalsize; for DiGCN, the teleport probability from \small{$\{0.05, 0.1, 0.15, 0.2\}$}\normalsize\ and $K$ from \small{$\{1, 2\}$}\normalsize
~\cite{digcn,zhang2021magnet};
for MagNet, $q$ in the magnetic Laplacian from \small{$\{0, 0.05, 0.1, 0.15, 0.2, 0.25\}$}\normalsize and $K$ from \small{$\{1, 2, 3\}$}\normalsize~\cite{zhang2021magnet}; etc.
For \ourseos, we tuned
$\lambda$ from \small{$\{0.01, 0.05, 1, 5\}$}\normalsize~and $K$ from \small{$\{1, 2, 3\}$}\normalsize.
For all GNNs, we used $2$ layers for a fair comparison. 
Models were optimized with Adam~\cite{kingma2014adam} following prior work~\cite{hyperbolicgat,hgcn,chami2020low}, with early stopping based on the validation results.

\section{Results}
\label{sec:result}

\noindent \textbf{Link Prediction.} 
We list the LP results of \ours in comparison to $10$ \textit{GNN} techniques using $4$ or $8$ dimensional node embeddings on Air and Cora in Table~\ref{tab:lp_result_fewdim}. One advantage of hyperbolic digraph embedding is low data distortion even with a low-dimensional embedding space.  The superior performance of \ours is evident---the highest relative gain of \ours is $\mathbf{21.43\%}$ on AP over the Cora dataset.
In addition, the difference from the mean
to the best metric value is considerably lower for \ours than other methods.
Given a low budget of embedding dimensionality, methods that use hyperbolic space (\ourseos, HAT and HGCN) are top performing, and the latest DRL GNNs (\ourseos, MagNet, DiGCN, and DGCN) overall outperform
traditional 
GNNs (GCN, VGAE, and GAT).

We report the LP performance of \ours in Table~\ref{tab:lp_result} in comparison to $14$ techniques by using a $32$-dimensional embedding space following the typical practice~\cite{digcn}. 
We can observe that techniques relying on matrix decomposition (ATP) or random walks (NERD, APP), are sensitive to outliers
and lack  
effectiveness and robustness. While standard deviations are omitted from the table due to space constraints, we have found that methods with higher average metric values typically have smaller standard deviations.
GNNs obtain higher scores.
Comparing Euclidean-based methods, DRL techniques
(marked with $^\text{\textdagger}$) 
can achieve better results than popular GNNs (e.g., GCN).
Still, methods that learn representations in hyperbolic space
(marked with $^\text{\textsection}$)
tend to be more 
competitive than those in Euclidean space.
With $32$-dimensional embeddings, gravity-augmented GCN and VGAE obtain better results than GCN and VGAE, and are able to occasionally hold the second or third position when ranking all $15$ methods based on their performance. As the dimensionality increases, the gap from \ours to the other methods decreases, but \ours remains the best-performing method across all datasets and metrics.

\noindent \textbf{Semi-supervised Node Classification.}  
Table~\ref{tab:nc_result} reports the NC results on CiteSeer and Cora-ML, and Table~\ref{tab:wiki_result} provides the results on Wiki.
\ourseos, which considers diverse neighborhoods with low distortion and is trained with self-supervision to preserve asymmetry, statistical significantly outperforms the state-of-the-art (SOTA) methods.
We increase the embedding dimensionality from $4$ up to $256$. The effectiveness of \ours is remarkable in low dimensionality regimes, yet \ours also remains the best method at a high dimensionality.
Unlike the LP task, DGCN and DiGCN often hold the second or third rank. However, due to sensitivity to tuned hyperparameters, their performance is unstable across dataset splits (i.e., occasionally extremely large standard deviations).

\begin{figure}[t]
	\centering
	\includegraphics[scale=0.30]{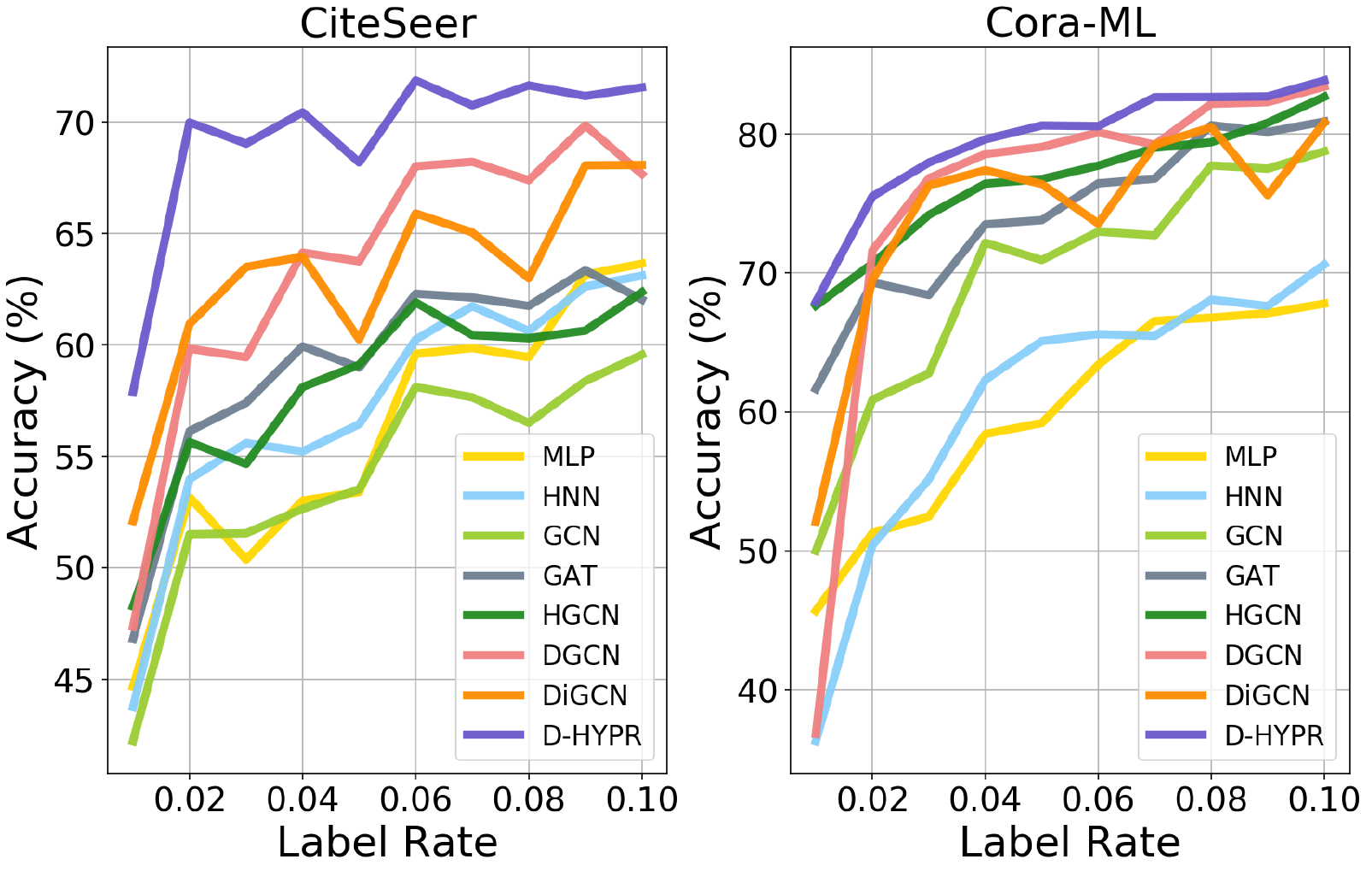}
	\vspace{-10pt}
	\caption{Accuracy on the \textit{Semi-supervised Node Classification} task by varying the ratio of nodes labeled for training. The embedding dimensionality is $32$.
	}
	\label{fig:vnc}
	\vspace{-5pt}
\end{figure}

We further follow prior work~\cite{node2vec} in reporting the results
when the number of nodes labeled for training is varied
between  $1\%$ and $10\%$.
According to Fig.~\ref{fig:vnc},  \ours \textit{consistently outperforms} the baselines,
and tends to perform well at fairly low label rates.

\noindent \textbf{Link Sign Prediction.}
Table~\ref{tab:wiki_result} reports the results of SP. 
\ours is the most effective GNN model.  Similar to LP and NC tasks, the effectiveness of \ours is the most striking using a $4$ dimensional embedding space.
One interesting observation is that the relative gains of \ours on Wiki NC is much higher than  Wiki SP, which suggests that asymmetry preservation can greatly improve the NC results (because unlike the NC task, while learning the asymmetric link sign prediction task, the baselines are able to simultaneously learn asymmetric node connectivity).

\noindent \textbf{Embedding Visualization.} 
In Fig.~\ref{fig:teaser}, we visualize $2$D projections of embeddings.
Unlike the prior methods (e.g., DGCN, DiGCN, etc.), in whose 2D projected embedding space 
nodes belonging to different topic classes often severely overlap, 
\ours leads to the best class separation. This suggests that \ours produces an embedding space that better captures the semantics of the digraph.

\noindent \textbf{Parameter Sensitivity.}
We first examine how $\lambda$ affects the performance of \ours by varying $\lambda$ from $0.25$ to $10$ (Table~\ref{tab:param_sen_lambda}) while keeping other hyperparameters fixed (32-Dim, the NC task).
    Larger $\lambda$ place
    more value on
    a symmetric embedding distance (which models `homophily'
    ), whereas a smaller $\lambda$ emphasizes the asymmetric node connectivity (which characterizes `preferential attachment'). The performance of \ours first increases with $\lambda$ and then decreases.
Using the $\lambda$ that produces the best result, we then vary $K$. As shown in Table~\ref{tab:param_sen_K},  better results are obtained when $K$  is larger, which
    means a wider receptive field and more scale information.
    However, an overly large $K$ can lead to feature dilution.  
    It is worth mentioning that \ours still outperforms the SOTA methods by a large margin when $K$=$1$, which suggests the superiority of \ours is not simply coming from the neighbor augmentation that connects nodes to their k-order neighbors. Hyperbolic neighborhood collaboration and preserving asymmetry are important factors that lead to the  superiority of \ourseos.

\noindent \textbf{Ablation Study.}
\label{sec:ablation}
As shown in Table~\ref{tab:ablation}, removing any neighborhood that we defined
harms the performance of \ourseos. 
Compared with the approach that learns the proximity matrices (adjacency matrix $A$ $+$ $3$ learnable matrices) or approaches that use other forms of multi-scale proximity matrices (e.g., MagNet, DiGCN and DGCN), \ours performs much better. The  proximity matrices are proposed in a way to leverage the inductive biases exhibited in real-world digraphs, thus facilitating the learning process and increasing the accuracy.
Replacing hyperbolic with the Euclidean space entails substantial performance drops.
    Still, this ablation yields better results than GCNs
    due to the other proposed components (e.g., collaborative learning).
Neighborhood collaboration is also crucial. The ablation that removes the hyperbolic neighborhood aggregation component has worse results than our full design, and the ablation that further replaces hyperbolic with the Euclidean space has much lower accuracies.  
Moreoever, self-supervision helps substantially. \ours is aided by the Gravity regularizer  more than the Fermi-Dirac regularizer,
    as the former captures
    the asymmetric link connectivity.
    While the Fermi-Dirac regularizer  provides auxiliary benefits, the embedding distance term in the Fermi-Dirac regularizer co-occurs in the Gravity regularizer, which also explains the stronger capability of the latter.
All ablations have a lower accuracy than our full model, suggesting that
the ablated components work together to increase the learning abilities of \ourseos.

\noindent \textbf{Discussion.}
\ours has a statistically superior and more stable performance across datasets and tasks. This is because \ours 
benefits from the use of hyperbolic space, information collected from the multi-ordered diverse neighborhoods and accounts for directionality.  
By favoring non-Euclidean over Euclidean geometry for DRL, \ours incurs lower node neighborhoods distortion.
In addition, the proposed $4$ canonical types of $k$-order proximity matrix are defined based on the semantics of directed edges in accordance with real-life observations. 
This allows
\ours to leverage inductive biases exhibited in many real-world digraphs, facilitating the learning and increasing the accuracy.

\begin{figure*}[t]
	\centering
	\vspace{-5pt}
	\includegraphics[width=\textwidth]{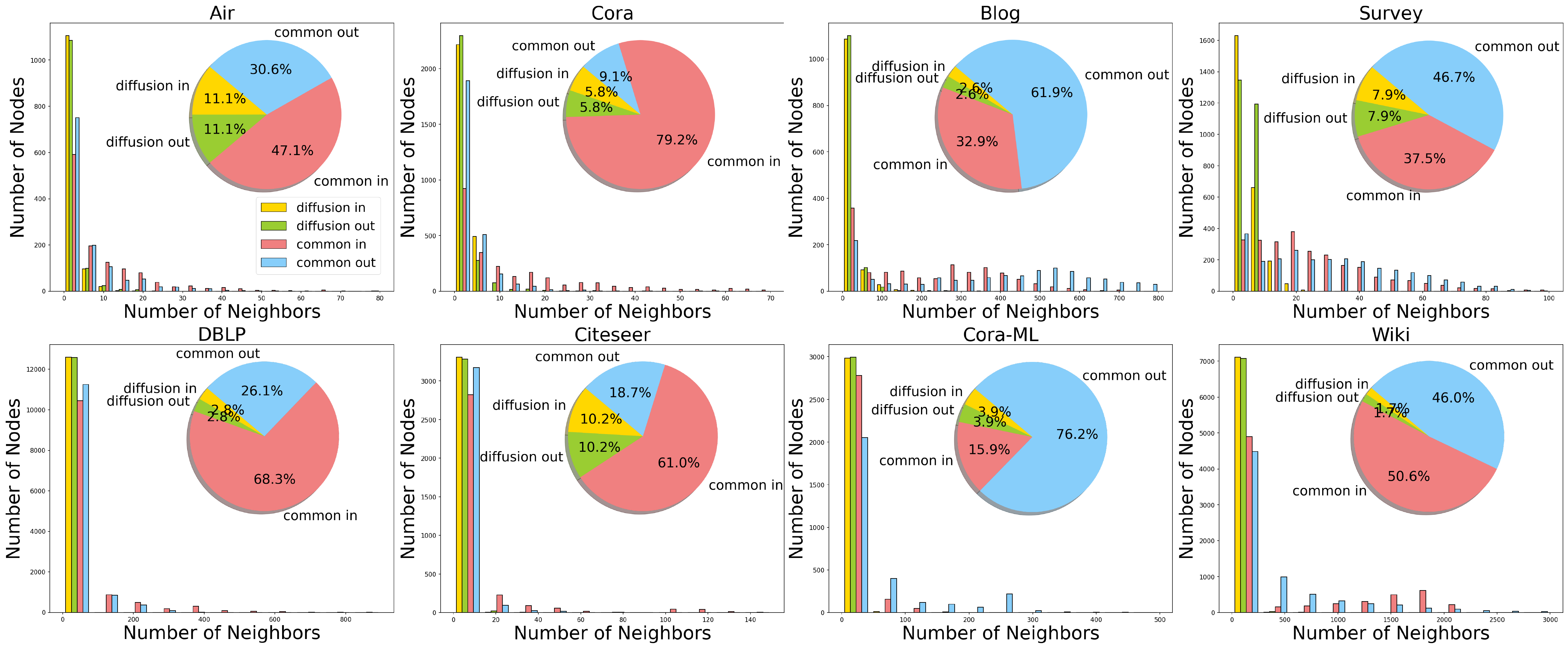}
	\vspace{-15pt}
	\caption{Neighborhood analyses of datasets. The \texttt{common in/out} neighborhood consists of more neighbors than \texttt{diffusion in/out} neighborhood that traditional methods typically use. The $8$ digraph datasets demonstrate a clear scale-free characteristic for most neighborhoods. }
	\label{fig:dataset}
	\vspace{-5pt}
\end{figure*}

Since \ours addresses \textit{Neighborhood Modeling}, we provide neighborhood analyses of datasets in Fig.~\ref{fig:dataset}, where pie charts show the ratio of
the $4$ canonical types of neighborhoods in each dataset ($K$=$1$).
Unlike the  \texttt{diffusion in/out} neighborhood that traditional 
GNNs typically use, \texttt{common in/out} neighborhood consists of more neighbors, which suggests that neighborhood collaborative learning benefits from encoding additional context.
Nevertheless, a larger neighborhood size does not necessarily entail a greater importance according to the ablation study (Table~\ref{tab:ablation}).
For each neighborhood type, we also plot a histogram showing the distribution of the number of neighbors a node has over the entire graph.
We observe 
asymptotical
power-law node-degree distributions 
(i.e., scale-free) for most neighborhoods in these digraph datasets.

\begin{table}[t]
\begin{center}
\caption{Experimental results 
on Wiki.
} 
\vspace{-10pt}
\label{tab:wiki_result}
\small
\fontsize{7.5pt}{7.5pt}\selectfont
\begin{tabular}{@{}llrr@{}}
\toprule
 &  \textbf{Model}     & \textbf{Node Classification} & \textbf{Link Sign Prediction} \\ \midrule
   \multirow{5}{*}{4-Dim} &  GCN~\cite{gcn}  & $17.01\pm0.1$     &  \magenta{\textit{78.96$\pm$0.4}}   \\ 
 &  GAT~\cite{gat}      &  \blue{\ul{40.75$\pm$10.7}} &  \blue{\ul{79.38$\pm$0.2}}   \\
 &  HGCN~\cite{hgcn}  &   \magenta{\textit{36.07$\pm$5.3}}  & $78.72\pm0.0$   \\
\cline{2-4}
 &  \ours (ours)     & \red{\textbf{$^\ast$71.27$\pm$0.79}}   &  \red{\textbf{$^\ast$79.83$\pm$0.0}}   \\
 &  Relative Gains ($\%$)    &  74.90  &  0.57  \\ \midrule
   \multirow{5}{*}{8-Dim} &  GCN~\cite{gcn}  &  $39.26\pm9.5$    &  $78.76\pm0.1$  \\ 
 &  GAT~\cite{gat}      & \magenta{\textit{46.78$\pm$ 10.5}}  &  \blue{\ul{79.41$\pm$0.2}}  \\
 &  HGCN~\cite{hgcn}  &  \blue{\ul{58.40$\pm$10.9}}    & \magenta{\textit{79.23$\pm$0.2}} \\
\cline{2-4} 
 &  \ours (ours)     &   \red{\textbf{$^\ast$70.53$\pm$1.6}}  &   \red{\textbf{$^\ast$79.47$\pm$0.3}}  \\ \cline{2-4}
 &  Relative Gains ($\%$)    &  20.77  &  0.08 \\ \midrule
  \multirow{5}{*}{32-Dim} &  GCN~\cite{gcn}  &   $37.77\pm 6.7$    & \magenta{\textit{79.39$\pm$ 0.1}} \\ 
 &  GAT~\cite{gat}      & \magenta{\textit{46.12$\pm$ 8.5}}  &   \blue{\ul{79.66$\pm$ 0.1}} \\
 &  HGCN~\cite{hgcn}  &   \blue{\ul{52.63$\pm$ 5.8}}  & $79.21\pm 0.2$ \\
\cline{2-4}
 &  \ours (ours)     &  \red{\textbf{$^\ast$71.65 $\pm$ 1.0}}    &     \red{\textbf{$^\ast$79.73 $\pm$ 0.2}}      \\ \cline{2-4}
 &  Relative Gains ($\%$)    &   36.14        & 0.09  \\   \bottomrule

\end{tabular}
\end{center}
\begin{flushleft}\small{Note: $10$ random dataset splits are used for the SP task. The embedding dimensionality is $32$. }\end{flushleft} 
\vspace{-5pt}
\end{table}

\begin{table*}[t]
\centering
\caption{Parameter sensitivity analysis in terms of  $\lambda$. The superiority of \ours is not sensitive to
the  hyperparameter $\lambda$. 
} 
\vspace{-8pt}
\label{tab:param_sen_lambda}
\setlength{\tabcolsep}{3pt}
\fontsize{7.3pt}{7.3pt}\selectfont
\begin{tabular}{@{}l|ccccccccccccccccccccc@{}}
\toprule
$\lambda$   & 0.25 & 0.50 & 0.75 & 1.00 & 1.25 & 1.50 & 1.75 & 2.00 & 2.25 & 2.50 & 2.75 & 3.00  & 3.25 & 3.50 & 3.75 & 4.00 & 4.25 & 4.50 & 4.75 & 5.00 & 10.0 \\ \midrule
\multirow{2}{*}{\textbf{CiteSeer}} & 69.74 & \textbf{70.66}  & 70.46  & 70.44 & 70.30  & 70.34  & 69.99  & 69.79  & 69.24  & 69.61 & 68.13  & 68.05  & 68.12  & 67.64  & 67.85  & 67.67 & 67.69  & 67.34 & 67.18  &  67.12 & 66.85\\
  &  $\pm$1.6  & $\pm$\textbf{1.2}  & $\pm$1.3   & $\pm$1.4  & $\pm$1.1  & $\pm$1.3  & $\pm$1.4  & $\pm$1.6  & $\pm$1.6 & $\pm$1.2 & $\pm$1.4  &  $\pm$1.3 & $\pm$1.9  & $\pm$1.8  & $\pm$1.8  & $\pm$1.9 & $\pm$1.9 & $\pm$2.3  & $\pm$2.1 &  $\pm$1.7 & $\pm$1.5  \\\midrule
\multirow{2}{*}{\textbf{Cora-ML}} & 81.29  & 81.18  & 81.59  & 81.68  & 81.83  & 81.97  & \textbf{82.16}  & 81.65  & 81.10 & 81.17 & 81.59 & 81.66  & 81.32  & 81.93  & 80.19 & 80.31 & 79.13  & 79.51 &  80.18 & 79.78 &  77.73  \\ 
  &  $\pm$1.3 & $\pm$1.2 & $\pm$1.2  & $\pm$1.4 & $\pm$1.1  & $\pm$1.0  & $\pm$\textbf{1.3}  & $\pm$1.2  & $\pm$1.0 & $\pm$1.4 & $\pm$1.0 & $\pm$1.2  & $\pm$ 1.1 & $\pm$1.1  & $\pm$1.5 & $\pm$1.3 & $\pm$2.3  & $\pm$1.7 &  $\pm$1.9 & $\pm$1.4 &  $\pm$2.0  \\  
 \bottomrule
\end{tabular}
\begin{flushleft}\small{Note: the task is Node Classification, and the embedding dimensionality is $32$ (same for Table~\ref{tab:param_sen_K}). }\end{flushleft} 
\end{table*}

\begin{table}[t]
\centering
\caption{Parameter sensitivity analysis in terms of  $K$. \ours consistently outperforms SOTA methods by a large margin.} 
\vspace{-8pt}
\label{tab:param_sen_K}
\fontsize{7.5pt}{7.5pt}\selectfont
\begin{tabular}{@{}l|ccc@{}}
\toprule
$K$   &  1 & 2 & 3 \\ \midrule
\textbf{CiteSeer} &  \text{69.23 $\pm$ 1.5}  & \textbf{70.66 $\pm$ 1.2} & \text{69.76 $\pm$ 1.5}  \\
\textbf{Cora-ML} &  \text{82.16 $\pm$ 1.3}  &  \text{82.16 $\pm$ 1.3}  & \textbf{82.19 $\pm$ 1.3}  \\
 \bottomrule
\end{tabular}
\end{table}

\begin{table}[t]
\centering
\caption{Ablation study that demonstrates the individual contribution of components in \ours (32-Dim, the NC task).
} 
\vspace{-8pt}
\label{tab:ablation}
\fontsize{7.5pt}{7.5pt}\selectfont
\begin{tabular}{@{}lcc@{}}
\toprule
\textbf{Method}      & \textbf{CiteSeer}  & \textbf{Cora-ML} \\ \midrule
\ours (Our Full Design)    &  $70.66 \pm 1.2$   &  $82.19 \pm 1.3$ \\ \midrule
No $A_{d_{in}}^{k}$        & $68.72\pm 1.2$  &  $82.11\pm 1.2$   \\
No $A_{d_{out}}^{k}$        &   $69.10\pm 0.9$   & $81.33\pm 1.4$ \\
No $A_{c_{in}}^{k}$        & $69.98\pm 1.0$  &  $81.86\pm 1.6$\\
No $A_{c_{out}}^{k}$     & $69.84\pm 1.3$  & $81.74\pm 1.8$  \\
No Hyperbolic Neighborhood Collaboration         &   $70.13\pm 1.5$  &   $82.03\pm 1.1$\\
No Gravity      & $68.58\pm 1.3$   &   $79.21\pm 1.5$  \\
No Fermi-Dirac       & $70.03\pm 1.2$   &  $82.05\pm 1.3$   \\
No Self-Supervision      &  $67.85\pm  1.9$   &  $78.15\pm 2.1$  \\\midrule
Euclidean      & 61.86 $\pm$ 5.4  &   73.38 $\pm$ 6.7   \\
Euclidean and No Neighborhood Collaboration    & $51.01 \pm 6.2$   & $65.46 \pm 12.1$  \\
\hline 
$A$ + Three Learnable Matrices
  & $60.97 \pm 12.7$  & $78.92 \pm 2.9$   \\ 
 \bottomrule
\end{tabular}
\vspace{-5pt}
\end{table}

Though in principle, MLP can serve as the node-pair score function to learn asymmetric node connectivity (e.g., used by MagNet, DiGCN, etc., in the LP experiments),
we resort to Fermi-Dirac and Gravity decoders because the two neatly model the 
driving forces of link formation,
and provide the right level of inductive biases for \ours to more easily generalize well across cases.
Fermi-Dirac is particularly suitable for hyperbolic geometry because Fermi-Dirac statistics provide a physical interpretation of hyperbolic distances as energies of links~\cite{krioukov2009curvature}, 
and the Gravity function is elegantly derived from Newton's theory of universal gravitation with the learnable mass 
encompassing centrality measures.
Overall, \ours leverages the inductive biases exhibited in real-world digraphs and thus generalizes well across tasks;
it  utilizes multi-ordered partitioned-neighborhoods with hyperbolic neighborhood collaboration to address \textit{Neighborhood Modeling},
and employs self-supervised learning with sociopsychology-inspired regularizers for \textit{Asymmetry Preservation}.

\section{Conclusion}
\label{sec:conclude}

We propose \ourseos: the \textbf{D}igraph \textbf{HYPER}bolic Network,
as a novel GNN-based formalism
for Digraph Representation Learning (DRL) by addressing \textit{Neighborhood Modeling} and \textit{Asymmetry Preservation}.
Through extensive and rigorous evaluation
involving \textit{\textbf{21}} prior techniques,
we empirically demonstrate the superiority
of \ourseos.
\ours outperforms the current SOTA consistently and statistically significantly on \textbf{\textit{8}} digraph datasets across \textit{\textbf{4}} tasks.
In addition, \ours retains effectiveness given a low budget of embedding dimensionality or labeled training samples,  which is desirable for
real-world applications.

One limitation of \ours is the increased number of parameters, due to the use of multiple neighborhoods.
As future work, we would like to explore automatic and dynamic neighborhood partitioning,
as well as
parameter-sharing mechanisms
to improve \ourseos.
Furthermore, theoretical analyses and novel large-scale applications of \ours 
are avenues worthy of exploration.

\subsubsection*{\textbf{Acknowledgment}}
The research was supported in part by NSF awards: IIS-1703883, IIS-1955404, IIS-1955365, RETTL-2119265, and EAGER-2122119.
This material is based upon work supported by the U.S. Department of Homeland Security under Grant Award Number 22STESE00001 01 01.

Disclaimer: The views and conclusions contained in this document are those of the authors and should not be interpreted as necessarily representing the official policies, either expressed or implied, of the U.S. Department of Homeland Security.

\bibliographystyle{ACM-Reference-Format}
\bibliography{reference}

\end{document}